\documentclass[10pt,twocolumn,letterpaper]{article}

\usepackage{cvpr}
\usepackage{times}
\usepackage{epsfig}
\usepackage{graphicx}
\usepackage{amsmath}
\usepackage{amssymb}

\usepackage[breaklinks=true,bookmarks=false]{hyperref}

\usepackage{tikz}
\usepackage{pgfplots}
\usepackage{makecell}
\usepackage{xr}
\usepackage{multirow}
\usepackage{color}
\usepackage{soul}

\usetikzlibrary{arrows}
\usetikzlibrary{shapes}
\usetikzlibrary{calc}
\usetikzlibrary{fit}
\usetikzlibrary{positioning}

\cvprfinalcopy % *** Uncomment this line for the final submission

\def\cvprPaperID{2486} % *** Enter the CVPR Paper ID here

\graphicspath{{.}{1_introduction/}{3_approach/}{4_experiments/}}

\ifcvprfinal\fi
\begin{document}

%%%%%%%%% TITLE
\title{Multi-Label Zero-Shot Learning with Structured Knowledge Graphs}
\author{Chung-Wei Lee${}^1$\footnotemark[1] , Wei Fang${}^1$\footnotemark[1] , Chih-Kuan Yeh${}^2$, Yu-Chiang Frank Wang${}^1$\\
Department of Electrical Engineering, National Taiwan University, Taipei, Taiwan${}^1$\\
Machine Learning Department, Carnegie Mellon University, Pittsburgh, USA${}^2$\\
{\tt \{b02901088,b02901054,ycwang\}@ntu.edu.tw, cjyeh@cs.cmu.edu}\\
}

\maketitle
\thispagestyle{empty}

\footnotetext[1]{Indicates equal contribution}

\begin{abstract}
In this paper, we propose a novel deep learning architecture for multi-label zero-shot learning (ML-ZSL), which is able to predict multiple unseen class labels for each input instance. Inspired by the way humans utilize semantic knowledge between objects of interests, we propose a framework that incorporates knowledge graphs for describing the relationships between multiple labels. Our model learns an information propagation mechanism from the semantic label space, which can be applied to model the interdependencies between seen and unseen class labels. With such investigation of structured knowledge graphs for visual reasoning, we show that our model can be applied for solving multi-label classification and ML-ZSL tasks. Compared to state-of-the-art approaches, comparable or improved performances can be achieved by our method.
\end{abstract}

\section{Introduction}
Real-world machine learning applications such as image annotation, music categorization, or medical diagnosis require assigning more than one class label to each input instance. Take image annotation for example, the learning models have to predict multiple labels like sky, sea, or ship for a single input image. 
Different from traditional multi-class methods which only predict one class label for each instance, learning multi-label classification models typically require additional efforts. More specifically, we not only need to relate the images with their multiple labels, it is often desirable to exploit label correlation due to the co-occurrences of the labels of interest.

In general, binary relevance~\cite{Tsoumakas07multi-labelclassification:} is the simplest solution to multi-label classification problems, which coverts the original task to multiple disjoint binary classification problems. However, it lacks the ability to model label co-occurrences, and thus might not be preferable. Approaches such as \cite{Read2011,icml2010_DembczynskiCH10} take cross-label correlation by assuming label priors, while label-embedding based methods \cite{BalasubramanianL12,doi:10.1162/NECO_a_00320,NIPS2012_4561,changpinyo2017predicting,changpinyo2016synthesized} project both input images and their labels onto a latent space to exploit label correlation. Methods that utilize deep neural networks have also been proposed. BP-MLL \cite{1683770} first proposed a loss function for modeling the dependency across labels, while other recent works proposed different loss functions \cite{DBLP:journals/corr/GongJLTI13,DBLP:journals/corr/NamKGF13} or architectures \cite{DBLP:journals/corr/WeiXHNDZY14,Wang_2016_CVPR,yeh2017learning} to further improve performance.

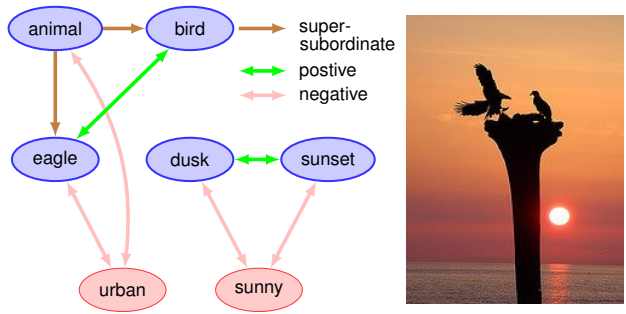
\begin{figure}[t]
\centering
    {\Large\sf
    \resizebox{\linewidth}{!}{%
    \begin{tikzpicture}
        \tikzstyle{state}=[ellipse,thick, minimum size=1.2cm, minimum width=2.4cm]
        \tikzstyle{node}=[circle,thick, minimum size=0.8cm]
        \tikzstyle{po}=[line width = 0.1cm, green,latex-latex]
        \tikzstyle{ne}=[line width = 0.1cm, pink,latex-latex]
        
        \tikzstyle{ncb}=[draw=blue!80, fill=blue!20]
        \tikzstyle{ncr}=[draw=red!80, fill=red!20]
        \tikzstyle{flow} = [draw,brown, line width = 0.1cm, -latex]
        
        \node (A) [state,ncb,ultra thick]{animal}; 
        \node (B) at (A)[state,ncb,right=2.5cm,ultra thick]{bird};
        \node (E) at (A) [state,ncb,below=3cm,ultra thick]{eagle};
        \node (D) at (E) [state,ncb,right=2.5cm,ultra thick]{dusk};
        \node (S) at (D) [state,ncb,right=2.5cm,ultra thick]{sunset};
        \node(EDM) at ($(E)!0.5!(D)$) {};
        \node(DSM) at ($(S)!0.5!(D)$) {};
        \node (U) at (EDM) [state,ncr,below=3cm]{urban};
        \node (sun) at (DSM) [state,ncr,below=3cm]{sunny};
        \draw[flow] (A)--(E);
        \draw[flow] (A)--(B);
        \draw[po] (D)--(S);
        \draw[ne] (U)to[bend right=20](A);
        \draw[ne] (U)--(E);
        \draw[ne] (D)--(sun);
        \draw[ne] (S)--(sun);
        \draw[po] (B)--(E);
        \node (hb) at (B)[right=1cm]{};
        \node (he) at (hb)[right=1.5cm]{};
        \node (hl) at (he)[right]{super-};
        \node (hf)[below right=-0.3cm and -1.77cm of hl]{subordinate};
        \draw[flow] (hb)--(he);
        
        \node (nb) at (hb)[below=1.0cm]{};
        \node (ne) at (nb)[right=1.5cm]{};
        \node (nl) at (ne)[right]{postive};
        \draw[po] (nb)--(ne);
        
        \node (pb) at (nb)[below=0.5cm]{};
        \node (pe) at (pb)[right=1.5cm]{};
        \node (pl) at (pe)[right]{negative};
        \draw[ne] (pb)--(pe);
        
        \node (fig) at (S)[right=2cm] {\includegraphics[height=8cm]{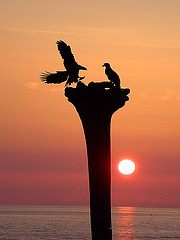}};
        
    \end{tikzpicture}
    }%
    }
    \caption{Illustration of structured knowledge graph for modeling the dependency between labels in the semantic space. We learn and utilize such graphs for relating the belief for each label, so that prediction of multiple seen or unseen labels can be achieved. The ground truth labels are noted in blue.}
    \label{fig:fig1}
\end{figure}

Extending from multi-label classification, multi-label zero-shot learning (ML-ZSL) is a branch of zero-shot learning (ZSL), which require the prediction of unseen labels which are not defined during training. Traditional multi-label approaches such as binary relevance or label-prior based methods obviously cannot be directly applied to ML-ZSL, since such methods lack the ability to generalize to unseen class labels. In contrast, approaches that utilize label representations in the semantic space such as label-embedding methods can be more easily adapted to ML-ZSL, given label representations of the unseen classes. Generally, label representations are obtained from human-annotated attribute vectors that describe the labels of interest either in a specific domain, or via distributed word embeddings learned from linguistic resources.

Nevertheless, although recent ML-ZSL methods such as \cite{Mensink_2014_CVPR,embedding2014ECCV,Zhang_2016_CVPR,Gaure_2017_UAI,Ren_etal_BMVC_17} have been proposed, existing approaches typically do not take advantages of structured knowledge and reasoning. Humans recognize objects not only by appearance, but also by using knowledge of the world learned through experience. Inspired by the above observation, we focus on leveraging existing structural knowledge for ML-ZSL, with the goal of deriving proper dependencies between different label concepts for both seen and unseen ones. Figure \ref{fig:fig1} illustrates how knowledge graphs can help in this problem, where we can model the co-occurring and non-co-occurring concepts and extend this knowledge to unseen classes with an external structured knowledge graph. There has been work on multi-label problems utilizing structured knowledge. 
\cite{deng2014large} introduced a graph representation that enforces certain relations between label concepts. 
\cite{hu2016learning} employed recurrent neural networks (RNN) \cite{hochreiter1997long,schuster1997bidirectional} to model positive and negative correlations between different concept layers. 
More recently, \cite{Marino_2017_CVPR} extended neural networks for graphs \cite{scarselli2009graph,DBLP:journals/corr/LiTBZ15} to efficiently learn a model that reasons about different types of relationships between class labels by propagating information in a knowledge graph. 

However, to the best of our knowledge, none of existing work advances structured knowledge reasoning for ML-ZSL. In this paper, we propose a novel ML-ZSL approach to observe and incorporate associated structured knowledge. Labels are represented with semantic vectors and an information propagation mechanism is learned from the label relations observed in the semantic space. The propagation of such label relation information is then used to modify the initial beliefs for each class label. Once the propagation process is complete, multi-label classification (or ML-ZSL) can be performed accordingly. Our model incorporates structured knowledge graphs observed from WordNet \cite{Miller:1995:WLD:219717.219748} into an end-to-end learning framework, while learning the label representations and information to be propagated in the semantic space. With this framework, we are able to achieve ZSL by assigning the unseen label embedding vector into our learning model. We will show the effectiveness of our model in advancing the structured knowledge for reasoning, which would benefit the task of ML-ZSL. 

The main contributions of this work are highlighted as follows:

\begin{itemize}
\item To the best of our knowledge, our model is among the first to advance structured information and knowledge graphs for ML-ZSL.
\item Our method advances a label propagation mechanism in the semantic space, enabling the reasoning of the learned model for predicting unseen labels.
\item With comparable performance on standard multi-label classification tasks, our method performs favorably against recent models for ML-ZSL.
\end{itemize}

%Our work utilizes semantic vectors to represent labels along with information propagation in knowledge graphs

\section{Related Work}
Remarkable developments on image classification has been observed over the past few years due to the availability of large-scale datasets like ImageNet \cite{deng2009imagenet} and the development of deep convolutional neural networks \cite{krizhevsky2012imagenet,He_2016_CVPR}. 

Among image classification tasks, multi-label classification  aims at predicting multiple labels for an input image, whcih can be achieved by the technique of binary relevance~\cite{Tsoumakas07multi-labelclassification:} using neural networks. To further improve the performance, label co-occurrence and relations between labels are considered in recent works. Label embedding methods are among the popular techniques, which transform labels into embedded label vectors, so that the correlation between labels can be exploited \cite{weston2011wsabie,doi:10.1162/NECO_a_00320,DBLP:journals/corr/GongJLTI13,lin2014multi,yeh2017learning}. As non-linear embedding approaches, deep neural networks have also been utilized for multi-label classification \cite{1683770,DBLP:journals/corr/NamKGF13,DBLP:journals/corr/GongJLTI13,DBLP:journals/corr/WeiXHNDZY14,Wang_2016_CVPR,yeh2017learning}.

Another way to determine the dependency between labels is via exploring explicit semantic relations between the labels. The Hierarchy and Exclusion (HEX) graph \cite{deng2014large} captures semantic relations: mutual exclusion, overlap and subsumption between any two labels, improving object classification by exploiting the label relations. The model is further extended to allow for soft or probabilistic relations between labels \cite{ding2015probabilistic}. Later, \cite{hu2016learning} introduced Structured Inference Neural Network (SINN). Inspired by the idea of Recurrent Neural Network (RNN) \cite{hochreiter1997long,schuster1997bidirectional}, positive correlation and negative correlation between labels are derived for bidirectionally propagating information between concept layers, which further improves the classification performance; {Focusing on single-label activity recognition, \cite{deng2016structure} view both activity of input image and actions of each person in that image as a graph, and utilize RNN to update the observed graph for activity prediction.} On the other hand, Graph Neural Networks \cite{scarselli2009graph}, \cite{DBLP:journals/corr/LiTBZ15} present architectures of Graph Gated Neural Networks (GGNN), which apply Gated Recurrent Units (GRU) \cite{04d503dffe9f4dcfbacf886d1dee56fb} and allow propagation on the graphs. As a modification of GGNN, Graph Search Neural Network (GSNN) \cite{Marino_2017_CVPR} is successfully applied for multi-label image classification to exploit explicit semantic relations in the form of structured knowledge graphs.

Different from multi-label classification, zero-shot learning (ZSL) is a challenging task, which needs to recognize test inputs as unseen categories. ZSL also attracts extensive attention from the vision community~\cite{NIPS2009_0395,akata2013label,socher2013zero,frome2013devise,NIPS2014_5290,norouzi14iclr,embedding2014ECCV,Lampert:2014:ACZ:2587733.2587824,akata2015evaluation,ba15iccv}, which is typically addressed by relating semantic information like attributes \cite{lampert2009learning,FarhadiEHF09} and word vectors \cite{NIPS2013_5021,pennington2014glove} to the presence of visual content.%\cite{socher2013zero} trains an unsupervised deep neural network model with knowledge about unseen visual categories from text corpora to obtain a rich image representation. Attribute Label Embedding \cite{akata2013label} applies joint embedding for attribute vectors and image features to outperform direct attribute prediction methods. Likewise, the Deep Visual-Semantic Embedding Model (DeViSE) \cite{frome2013devise} explicitly projects images into a semantic embedding space, and hence can predict classes which have never been observed. 

\begin{figure*}[t]
	\centering
    \begin{tikzpicture}
        \tikzstyle{state}=[ellipse,thick, minimum size=0.6cm, minimum width=1.6cm]
        \tikzstyle{node}=[circle,thick, minimum size=0.8cm]
        \tikzstyle{output}=[thick, minimum size=0.4cm]
        \tikzstyle{input}=[thick, minimum height=4cm, draw=black!80, fill=black!20]
        
        \tikzstyle{ncb}=[draw=blue!80, fill=blue!20]
        \tikzstyle{ncg}=[draw=green!80, fill=green!20]
        \tikzstyle{ncr}=[draw=red!80, fill=red!20]
        \tikzstyle{nco}=[draw=orange!80, fill=orange!20]
        \tikzstyle{ndis}=[below=0.6cm]
        \tikzstyle{ndis2}=[below=0.4cm]
        \tikzstyle{ndis3}=[right=2.8cm]
        \tikzstyle{ndis4}=[right=2.3cm]
        \tikzstyle{ndisa}=[above=0.6cm]
        \tikzstyle{edg}=[very thick, latex-latex]
        \tikzstyle{bnd} = [draw, very thick, dashed, inner sep=0.5cm, black!66]
        \tikzstyle{flow} = [draw, thick, -latex, black]

        \newcommand{\hz}[1]{{\mathbf{h}^{(0)}_{#1}}}
        \newcommand{\hm}[1]{{\mathbf{h}^{(t)}_{#1}}}
        \newcommand{\hT}[1]{{\mathbf{h}^{(T)}_{#1}}}
        
        \newcommand{\dis}{2.5cm}
        
        \node (h1) [state,ncb]{$\hz{1}$};
        \node (h2)at(h1) [state,ncg,ndis]{$\hz{2}$};
        \node (h3)at(h2) [state,ncr,ndis]{$\hz{3}$};
        \node (dot)at(h3) [ndis2]{$\vdots$};
        \node (h4)at(dot) [state,nco,ndis]{$\hz{|\mathcal{S}|}$};
        
        \node (in) at (h3) [left = 2.5cm, input]{$\mathbf{x}$};
        
        \node[inner sep=0,outer sep=0] (sp) at ($(in)!0.3!(h3)$) {};
        \draw[flow] (in)--(sp)|-(h4);
        \draw[flow] (in)--(h3);
        \draw[flow] (in)--(sp)|-(h2);
        \draw[flow] (in)--(sp)|-(h1);
        \node (fi) [below right=0.2cm and -0.08cm of sp]{\large $\mathbf{F}_I$};
        
        \node (n2) at (h3) [node,ncg,right=\dis]{$\hm{2}$};
        \node (n1) [node,ncb,above right=1.4cm and 0.4cm of n2]{$\hm{1}$};
        \node (n3) [node,ncr,below right=1.2cm and 0.4cm of n2]{$\hm{3}$};
        \node (n4) at (n2) [node,nco,right=3cm]{$\hm{|\mathcal{S}|}$};
        \node[rotate=-25] (na) at ($(n1)!0.5!(n4)$) {\huge $ \ddots$};
        \node[rotate=65] (nb) at ($(n3)!0.5!(n4)$) {\huge $ \ddots$};
        \draw[edg](n1)--(n2);
        \draw[edg](n2)--(n4);
        \draw[edg](n2)--(n3);
        \draw[edg](n3)to[bend right=30](n4);
        
        \node (f3) at (n4) [state,ncr,ndis4]{$\hT{3}$};
        \node (f2) at (f3) [state,ncg,ndisa]{$\hT{2}$};
        \node (f1) at (f2) [state,ncb,ndisa]{$\hT{1}$};      
        \node (fd) at (f3) [ndis2]{$\vdots$};
        \node (f4) at (fd) [state,nco,ndis]{$\hT{|\mathcal{S}|}$};
        
        \node (o1)at(f1) [output,ncb,ndis3]{$1$};
        \node (o2)at(f2) [output,ncg,ndis3]{$0$};
        \node (o3)at(f3) [output,ncr,ndis3]{$1$};
        \node (od)at(fd) [ndis3]{$\vdots$};
        \node (o4)at(f4) [output,nco,ndis3]{$0$};
        
        \node(init) [bnd, fit=(h1) (h4)] {};
        \node(prop) [bnd, fit=(n1) (n2) (n3) (n4)] {};
        \node(finl) [bnd, fit=(f1) (f4)] {};
        \node(pred) [bnd, fit={($(o1)+(-0.3cm,+0.5cm)$) ($(o4)+(0.3cm,-0.5cm)$)}] {};
        
        \draw[flow] (init)--(prop){};
        \draw[flow] (prop)--(finl){};
        
        \draw[flow] (f1)--(o1);
        \draw[flow] (f2)--(o2);
        \draw[flow] (f3)--(o3);
        \draw[flow] (f4)--(o4);
        
        \node (fo) [below right=-0.1cm and 0.8cm of f3]{\large $\mathbf{F}_O$};
        
    \end{tikzpicture}
    
	\caption{Illustration of structured graph propagation for multi-label classification. Given an input $\mathbf{x}$, we calculate the initial belief state $\mathbf{h}_v^{(0)}$ for each label node. The resulting information is propagated via the observed graph for updating the associated belief states. After propagating $T$ times, the final belief states can be obtained for predicting the final multi-label outputs.}\label{fig:fig2}
\end{figure*}
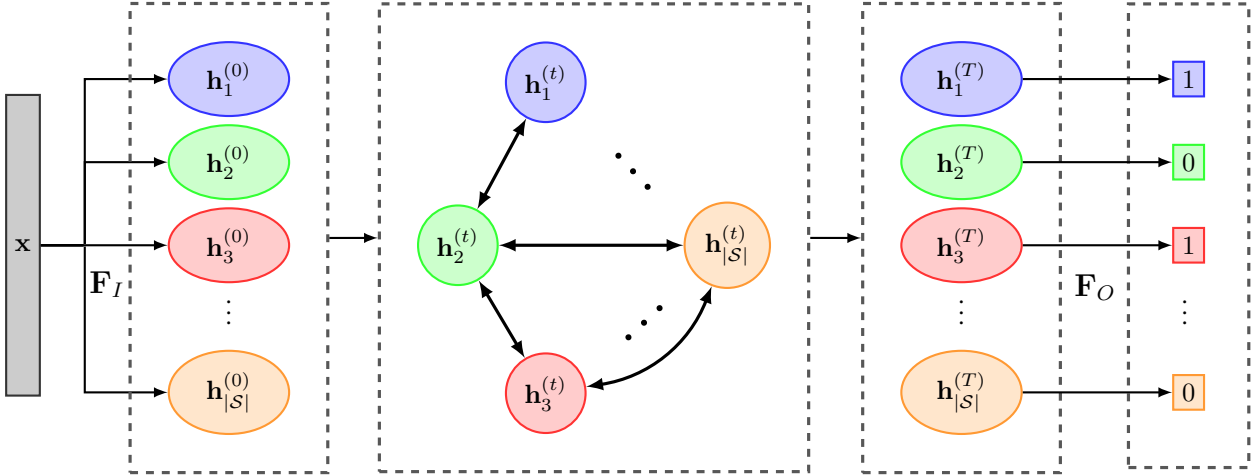

Extended from ZSL, multi-label zero-shot learning (ML-ZSL) further requires one to assign multiple unseen labels for each instance. To solve ML-ZSL tasks, COSTA\cite{Mensink_2014_CVPR} assumes co-occurrence statistics and estimates classifiers for seen labels by weighted combinations of seen classes. \cite{embedding2014ECCV} achieves ML-ZSL by exhaustively listing all possible combinations of labels and treating it as a zero-shot classification problem. Recently, \cite{Zhang_2016_CVPR} considers the separability of relevant and irrelevant tags, proposing a model
that learns principal directions for images in the embedding space. Multiple Instance Visual-Semantic Embedding (MIVSE) \cite{Ren_etal_BMVC_17} is another joint embedding method, which uses a region-proposal method to discover meaningful subregions in images and then maps the subregions to their corresponding labels in the semantic embedding space. \cite{Gaure_2017_UAI} leverages co-occurrence statistics of seen and unseen labels and learns a graphical model that jointly models the label matrix and the co-occurrence matrix.

\section{Our Proposed Approach}
\subsection{Notations and Overview}
We first define the notations used in this paper. Let $\mathcal{D}=\{(\mathbf{x}^i,\mathbf{y}^i)\}_{i=1}^N$ denote the set of training instances, where $\mathbf{x}^i \in \mathbb{R}^{d_{feat}}$ are $d_{feat}$-dimensional features and $\mathbf{y}^i \in \{0,1\}^{|\mathcal{S}|}$ are the corresponding labels in the label set $\mathcal{S}$.
Note that $N$ denotes the number of training instances, while $|\mathcal{S}|$ is the number of seen labels. 
Given $\mathcal{D}$ and $\mathcal{S}$, the task of multi-label classification is to learn a model such that the label $\hat{\mathbf{y}}\in \{0,1\}^{|\mathcal{S}|}$ of a test instance $\hat{\mathbf{x}}\in \mathbb{R}^{d_{feat}}$ can be predicted accurately.

For ML-ZSL, we have the unseen label set as $\mathcal{U}$, and the goal is to predict the labels in both $\mathcal{S}$ and $\mathcal{U}$ for a test instance $\hat{\mathbf{x}}$.
The predicted label is as $\tilde{\mathbf{y}}\in \{0,1\}^{|\mathcal{S}|+\mathcal{|U|}}$, where the first $|\mathcal{S}|$ dimensions are the predictions for the seen label set $\mathcal{S}$, and the bottom $|\mathcal{U}|$ dimensions are for the unseen ones. 

Since the images are without the annotation of labels $\mathcal{U}$ during training, ML-ZSL needs to extract the semantic information from the observed label space. In our proposed model, we use distributed word embeddings to represent a class label with a semantic vector. The word embedding is denoted as $\mathbf{W} = \{\mathbf{\mathbf{w}}_v\}_{v=1}^{|\mathcal{S}|+\mathcal{|U|}}$, where $\mathbf{w}_v \in \mathbb{R}^{d_{emb}}$ is the word vector representation for label $v$ in $\mathcal{S}\,\cup\,\mathcal{U}$, and $d_{emb}$ is the dimension of the word embedding space. 
In our work, we utilize GloVe \cite{pennington2014glove} as $\mathbf{W}$ with $d_{emb}=300$.

Our approach is illustrated in Figure~\ref{fig:fig2}. We take every label as a node with states in our structured knowledge graph. The initial belief states of these nodes $\mathbf{h}_v^{(0)}$ are first obtained through the input function $\mathbf{F}_I$, and the resulting information is propagated via the structured knowledge graph for updating the belief states. The propagation mechanism from each label node $u$ to a connecting node $v$ is governed by propagation weights $\mathbf{a}_{vu}$, which are produced from the relation function $\mathbf{F}^k_R$. We note that, this relation function takes the label representations $\mathbf{w}_u$ and $\mathbf{w}_v$ as inputs, where $k$ denotes the type of relation between nodes $u$ and $v$ as defined in the knowledge graph. The above propagation and interaction process would terminate after $T$ steps, followed by passing through a output function $\mathbf{F}_O$ to produce the final classification probabilities. In the following subsections, we will give details of how this model is used for ML-ZSL.

\subsection{Structured Knowledge Graph Propagation in Neural Networks}
Inspired by Graph Gated Neural Networks \cite{DBLP:journals/corr/LiTBZ15,Marino_2017_CVPR}, we consider a graph with $|\mathcal{S}|$ nodes, and the propagation model is learned with a gated recurrent update mechanism which is similar to recurrent neural networks. For the task of ML-ZSL, each node $v$ in the graph corresponds to a class label, and there is a belief state vector $\mathbf{h}_v^{(t)} \in \mathbb{R}^{d_{hid}}$ at every time step $t$. Following \cite{Marino_2017_CVPR}, we set $d_{hid}$ to 5. For ML-ZSL, we cannot simply apply an existing detector as in GSNN to obtain the initial belief states. Instead, we utilize an input function $\mathbf{F}_I(\mathbf{x},\mathbf{w}_v)$ that takes the input feature $\mathbf{x}$ and the label representation $\mathbf{w}_v$ for each node $v$ as inputs to calculate the initial belief state $\mathbf{h}_v^{(0)}$.
The function $\mathbf{F}_I$ is implemented by a neural network.

Next, using the structure of the knowledge graph which encodes the propagation weight matrix $\mathbf{A} \in \mathbb{R}^{|\mathcal{S}|d_{hid}\times |\mathcal{S}|d_{hid}}$, we retrieve the belief states of adjacent nodes and combine the information from adjacent nodes to get an update vector $\mathbf{u}_v^{(t)}$ for each node. The belief states are then updated by a gating mechanism by Gated Recurrent Unit (GRU) with $\mathbf{u}_v^{(t)}$ as the input. 

For each class label node $v \in \mathcal{S}$, the propagation recurrence is as follows:
\begin{flalign}
	\mathbf{h}_v^{(0)} &= \mathbf{F}_I(\mathbf{x}, \mathbf{w}_v), \label{eq:1}\\
    \mathbf{u}_v^{(t)} &= tanh\big(\mathbf{A}_v^\top 
    	\big[\mathbf{h}_1^{(t-1)}{}^\top\dots\mathbf{h}_{|\mathcal{S}|}^{(t-1)}{}^\top\big]^\top\big), \label{eq:2}\\
    \mathbf{h}_v^{(t)} &= GRUCell\big(\mathbf{u}_v^{(t)}, \mathbf{h}_v^{(t-1)}\big),\label{eq:3}
\end{flalign}
where $\mathbf{A}_v \in \mathbb{R}^{|\mathcal{S}|d_{hid}\times d_{hid}}$ is a submatrix of $\mathbf{A}$ that represents the propagation weight matrix for node $v$ (as detailed in the next subsection). $GRUCell$ is the GRU update mechanism, which is defined as:
\begin{flalign}
	\mathbf{z}_v^{(t)} &= \sigma\big(
    	\mathbf{W}^z \mathbf{u}_v^{(t)} + \mathbf{U}^z \mathbf{h}_v^{(t-1)} + \mathbf{b}^z
        \big), \label{eq:4}\\
	\mathbf{r}_v^{(t)} &= \sigma\big(
    	\mathbf{W}^r \mathbf{u}_v^{(t)} + \mathbf{U}^r \mathbf{h}_v^{(t-1)} + \mathbf{b}^r
        \big), \label{eq:5}\\
    \tilde{\mathbf{h}}_v^{(t)} &= tanh\big(
    	\mathbf{W}^h \mathbf{u}_v^{(t)} + \mathbf{U}^h(\mathbf{r}_v^{(t-1)}\odot \mathbf{h}_v^{(t-1)}) + \mathbf{b}^h
    	\big), \label{eq:6}\\
    \mathbf{h}_v^{(t)} &= (1-\mathbf{z}_v^{(t)})\odot \mathbf{h}_v^{(t-1)}+\mathbf{z}_v^{(t)}\odot \tilde{\mathbf{h}}_v^{(t)}, \label{eq:7}
\end{flalign}
where $\mathbf{W}$, $\mathbf{U}$, and $\mathbf{b}$ are learned parameters.

For each time step $t$, the confidence for each label node is obtained by the output function $\mathbf{F}_O$:
\begin{flalign}
	p_v^{(t)} = \mathbf{F}_O(\mathbf{h}_v^{(t)}),
\end{flalign}
which is implemented by a standard fully-connected neural network. After $T$ time steps for propagation, the final confidences $p_v^{(T)}$ would be obtained.

\subsection{Learning of the Propagation Matrix}
With the gated update mechanism for updating the belief state of each node in a graph, we now address a critical issue that how our model reasons and combines information from adjacent nodes lies in the matrix $\mathbf{A}_v$.

In~\eqref{eq:2}, we see that the update vector $\mathbf{u}_v^{(t)}$ is a weighted combination of the belief states of all other nodes by the propagation parameters in $\mathbf{A}_v$, with each hidden dimension having its own weights.
By constraining $\mathbf{A}_v$ to have non-zero weights for the elements that correspond to adjacent nodes and setting weights for non-adjacent nodes to zero, a node would combine information from only relevant nodes that are defined in the structured knowledge graph to obtain the update vector $\mathbf{u}_v^{(t)}$ for updating its own belief state.

In GSNN, the structured knowledge graph is defined with around 30 relation types. While the elements in $\mathbf{A}$ are learned, the edges of the same relation type are fixed in GSNN. This might limit its practical uses due to only a few relation types can be determined beforehand. In ML-ZSL, it is desirable to exploit finer relation between labels, so the propagation mechanism with the resulting knowledge graph would be sufficiently informative.

\begin{figure}[t]
	%\begin{tikzpicture}
	%	\node (test) [rectangle, minimum width=\linewidth, minimum height=3cm, draw=black] {};
	%\end{tikzpicture}
    \includegraphics[width=\linewidth]{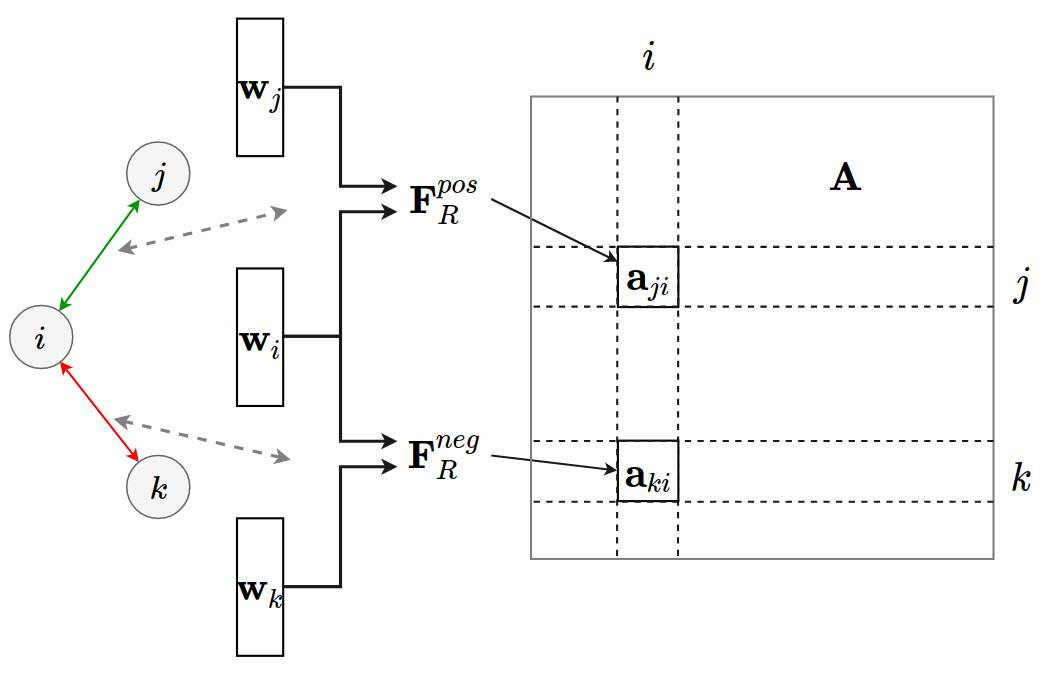}
    \caption{Learning of propagation matrix $\mathbf{A}$ in the semantic space via relation functions $\mathbf{F}^k_R$, with edges defined by the knowledge graph. Note that we only show the propagation from node $i$ outwards, but the matrix $\mathbf{A}$ would be symmetric in practice.}
    \label{fig:fig3}
\end{figure}

\begin{figure*}[t]
\includegraphics[width=\linewidth]{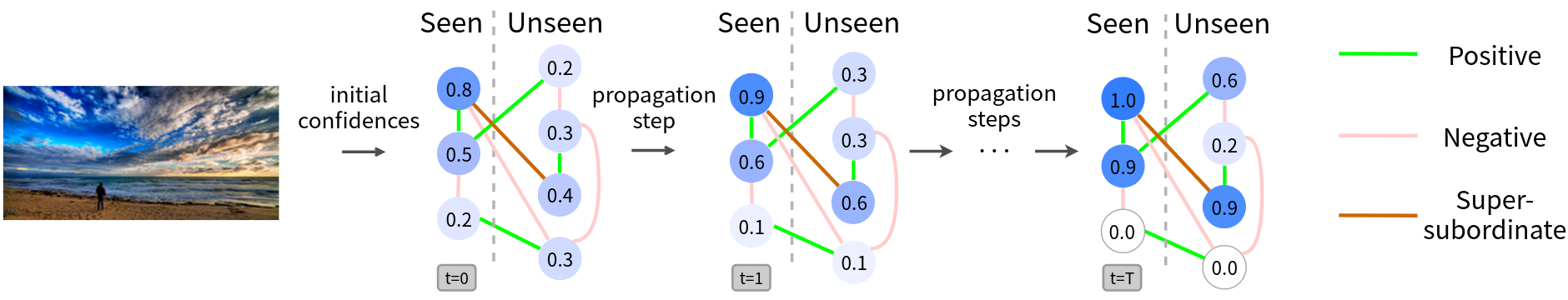}
\caption{Illustration of information propagation in the knowledge graph. Note that information from a belief state interacts with its adjacent seen/unseen states at each time step based on the relation observed in the knowledge graph. The probabilities represent the outputs of $\mathbf{F}_O$ applied to the belief states at each time step for illustration purposes.}
\label{fig:fig4}
\end{figure*}

To address the above concern, we propose a unique strategy as the propagation weight learning scheme. We combine the informations of word vector into knowledge graphs during propagation stages. Our scheme shares the propagation mechanism for the same relation types while being preferable for ZSL and other practical applications. More precisely, instead of assigning the same propagation weights for edges of the same type/relation, we alternatively assign the same relation function $\mathbf{F}_R^k$ that produces the propagation weights, where $k$ denotes the edge type. 
Given an edge in edges $\mathcal{E}$ that has edge type $k$, the propagation weights $\mathbf{a}_{vu} \in \mathbb{R}^{d_{hid}\times d_{hid}}$ are determined by:
\begin{flalign}
	\mathbf{a}_{vu} = \mathbf{F}_R^k(\mathbf{w}_v, \mathbf{w}_u),
\end{flalign}
where $\mathbf{w}_v$ and $\mathbf{w}_u$ are the word vectors for the class label nodes $v$ and $u$. The mechanism for learning propagation weights is illustrated in Figure \ref{fig:fig3}, in which each element of the matrix $\mathbf{a}_{vu}$ is determined by a unique bilinear form from joint embedding of the two associated labels. This allows our model to properly describe relationships between different nodes/relations. %Furthermore, we normalized each column of $\mathbf{A}$ and found this step to be crucial to learning the propagation mechanism, since the unnormalized propagation matrix would result in saturated update vectors $\mathbf{u}_v^{(t)}$.

As a final remark, for each edge type $k$, the function $\mathbf{F}_R^k$ learns a mapping from the semantic word embedding space to the propagation weights, so that the dependency between such relation edges can be modeled accordingly. More importantly, learning from the semantic space allows the aforementioned model to generalize to unseen class labels. Thus, the proposed scheme using relation functions $\mathbf{F}_R^k$ to determine the propagation weights $\mathbf{a}_{vu}$ would be especially preferable for ML-ZSL.

\subsection{From ML to ML-ZSL}
During training, the propagation weight matrix $\mathbf{A}$ can be obtained by forward passing through the relation networks $\mathbf{F}_R^k$, and is then used for information propagation described in~\eqref{eq:2} to \eqref{eq:7}. The loss function of our model is a weighted sum of the binary cross-entropy (BCE) of each label node, after the output of network $\mathbf{F}_O$ is observed at each time step. To be more precise, the loss $\mathcal{L}$ is defined as:
\begin{flalign}
\displaystyle
\mathcal{L}=\frac{1}{N}\frac{1}{|\mathcal{S}|}
\sum_{i, v, t}
\alpha(t)\Big((y^i_v\log p_v^{(t)}+(1-y^i_v)\log(1-p_v^{(t)})\Big),
\end{flalign}
where the weights $\alpha(t)=1/(T-t+1)$ encourage accurate predictions as $t$ increases. During the inference stage of multi-label classification, the final confidences $p_v^{(T)}$ at time step $T$ are used as the predicted outputs.

For ML-ZSL prediction, we extend $\mathbf{A}$ to $\tilde{\mathbf{A}}\in \mathbb{R}^{(|\mathcal{S}|+|\mathcal{U}|)d_{hid}\times(|\mathcal{S}|+|\mathcal{U}|)d_{hid}}$, so that it would encode relations of unseen class labels in the constructed knowledge graph.
We also constrain $\tilde{\mathbf{A}}$ so that for edges between $
\mathcal{S}$ and $\mathcal{U}$ we only allow propagation from seen to unseen nodes.
The update vector $\mathbf{u}_v^{(t)}$ is then calculated from the adjacent nodes for both seen classes $\mathcal{S}$ and unseen classes $\mathcal{U}$. Thus, we have~\eqref{eq:2} modified as:
\begin{flalign}
    \mathbf{u}_v^{(t)} &= tanh\big(\tilde{\mathbf{A}}_v^\top 
    	\big[\mathbf{h}_1^{(t-1)}{}^\top\dots\mathbf{h}_{(|\mathcal{S}|+|\mathcal{U}|)}^{(t-1)}{}^\top\big]^\top\big), 
        \forall v \in \mathcal{S} \cup \mathcal{U}.
\end{flalign}

The above model is able to calculate the initial belief states for the unseen class labels with $\mathbf{F}_I$, and performs propagation from seen to unseen labels (and also between unseen labels with $\tilde{\mathbf{A}}$ obtained through $\mathbf{F}_R^k$). Finally, the output confidence for each unseen label is derived by $\mathbf{F}_O$.
An illustration of the propagation mechanism for ML-ZSL is shown in Figure \ref{fig:fig4}, where the model generalizes from its initial beliefs on seen nodes to the unseen nodes.
We note that, during ML-ZSL, our model is also able to produce predictions for the seen class labels in addition to the unseen class labels. Thus, it can be considered for the more challenging task of generalized ML-ZSL.

\section{Experiments}

\subsection{Building the Knowledge Graph}
Before presenting the experimental results, we detail how we built the structured knowledge graph in our model. In our work, we consider WordNet \cite{Miller:1995:WLD:219717.219748} as the source for constructing the knowledge graph, since it is easily accessible and contains rich semantic relationships between different concepts.

We defined $3$ types of label relations for the knowledge graph: \textit{super-subordinate}, \textit{positive correlation}, and \textit{negative correlation}.
Super-subordinate correlations, also called hyponymy, hypernomy, or ISA relation, is defined and can be directly extracted from WordNet. For positive and negative relations between class labels, label similarities are calculated by WUP similarity \cite{wu1994verbs}, followed by thresholding the soft similarities into positive and negative correlations. As for label pairs with similarities between the positive and negative thresholds, or pairs without similarities from WUP similarity, they are viewed as not having any direct relation between them.

In addition, if a pair of labels exhibit super-subordinate relation, we directly apply its resulting dependency in our graph and do not further calculate its positive/negative relation. In the following experiments, we fix the propagation steps on the structured knowledge graph to 5 ($T=5$).

\subsection{Datasets and Settings}
To evaluate the performance of our model, we consider the following datasets for experiments: NUS-WIDE \cite{chua2009nus} and Microsoft COCO \cite{502}. For the multi-label classification task we perform experiments on both datasets, while NUS-WIDE is particularly applied for ML-ZSL evaluation. 

NUS-WIDE is a web image dataset including 269,648 images and the associated tags from Flickr. For these images, it consists of 1000 noisy labels collected from the web with 81 dedicated ground-truth concepts. We denote these two sets of labels as NUS-1000 and NUS-81, respectively. After collecting all existing images and removing images that do not have any tags, we obtain 90,360 images. We extract 2048-dimensional ResNet-152 \cite{He_2016_CVPR} feature representations from the images and use them as inputs for the following tasks. We further split the dataset into 75,000 training images, 5,000 validation images and 10,360 test images.

Microsoft COCO (MS-COCO) is a large-scale dataset for object detection, segmentation, and image captioning. We follow the 2014 challenge for data split (i.e., 82783 and 40504 images for training and testing, respectively) with 80 distinct object tags. After removing images without any labels, we split the training set into 78081 training images and 4000 validation images, and the test set is with 40137 images. For all the methods considered in our experiments, we extract and fix 2048-dimensional image features are extracted from ResNet-152.

\subsection{Multi-Label Classification}
We fist consider the conventional multi-label classification tasks for evaluating our proposed model.
For comparison, we consider WSABIE \cite{weston2011wsabie}, WARP \cite{DBLP:journals/corr/GongJLTI13}, and logistic regression (all with the above CNN features) as baseline approaches. We also implement Fast0Tag \cite{Zhang_2016_CVPR} to compare against models that are designed to handle multi-label classification problems (and later the ML-ZSL tasks).%, and DNN-BCE (DNN using binary cross-entropy loss). 

For testing, since WSABIE, WARP and Fast0Tag predict labels according to the ranking scores of the tags, we choose the top $K$ labels. Following conventional settings, we report results for $K=3$. As for logistics and our model, every label reports a final confidence for evaluation. Using the validation set, we select a proper probability threshold for predicting labels. Finally,
the metrics of precision (P), recall (R) and F1-measure are considered, which are commonly used in previous work.

\begin{table}
\small
\centering
  	\begin{tabular}{|c|ccc|ccc|}
    	%\Xhline{2\arrayrulewidth}
           \multirow{3}{*}{Method}& \multicolumn{3}{c|}{NUS-81} & \multicolumn{3}{c|}{MS-COCO} \\ 
           \cline{2-7}
           & P & R & F1 & P & R & F1 \\
          \hline
          WSABIE & 30.7& 52.0& 38.6 & 59.3 & 61.3 & 60.3   \\
          WARP & 31.4 & 53.3 & 39.5 &  60.2 & 62.2 & 61.2 \\
          Logistics &41.9& 46.2& 43.9 & 70.8 & 63.3 & 66.9  \\  
          Fast0Tag & 31.9& 54.0& 40.1 & 60.2 & 62.2 & 61.2 \\
          Ours & 43.4& 48.2 & {\bf 45.7} & 74.1 & 64.5 & {\bf 69.0} \\
        %\Xhline{2\arrayrulewidth}
      \end{tabular}
      \vspace{1.5em}
      \caption{Multi-label classification results on NUS-WIDE with 81 labels and MS-COCO with 80 labels. Results for WSABIE, WARP and Fast0Tag are with $K=3$.}
      \label{tb:ml}
\end{table}

\begin{table}[t]
	\small
	\centering
  	\begin{tabular}{|c|ccc|ccc|}
        %\Xhline{2\arrayrulewidth}
           \multirow{3}{*}{Method}& \multicolumn{3}{c|}{ML-ZSL} & \multicolumn{3}{c|}{Generalized} \\
           \cline{2-7}
           & P & R & F1 & P & R & F1\\
          \hline
           Fast0Tag ($K=3)$& 21.7 & 37.7& 27.2 & -& -& - \\
          Fast0Tag ($K=10$)& - & - & -&19.5 & 24.9&21.9 \\
          Ours w/o Prop. & 31.8& 25.1& 28.1& 24.3& 23.4& 23.9 \\
          %Ours w/o Prop. & 28.9& 22.1& 25.1& 31.6& 19.3& 24.0 & 21.9& 22.3& 22.1& 13.7& 10.9& 12.1 \\
          Ours &29.3& 31.9 & {\bf 30.6} & 22.8& 25.9 & {\bf 24.2} \\
        %\Xhline{2\arrayrulewidth}
          
      \end{tabular}
      \vspace{1.5em}
      \caption{Results for the ML-ZSL and generalized ML-ZSL tasks on NUS-1000 with 81 unseen labels and 925 seen labels.}
      \label{tb:mlzsl}
\end{table}

Table~\ref{tb:ml} lists and compares the results for the NUS-81 and MS-COCO datasets. We can see that our model produced comparable performances against baselines. It is worth noting that, since our model is not explicitly designed for solving multi-label but zero-shot learning, the above results sufficiently support the use of our model for multi-label classification. In addition, compared to Fast0Tag, which is designed for ML-ZSL and can also be used in the conventional multi-label setting, our model clearly achieved improved results on both datasets.

We also note that, although Fast0Tag reported higher scores on the recalls on NUS-81, it was not able to produce satisfactory results on the precisions.
The discrepancy between precision and recall can also be observed from the results in~\cite{Zhang_2016_CVPR}.
Similar remarks can be made for both WSABIE and WARP baselines. A possible explanation is that the number of tags in an image varies across the dataset, and thus simply choosing the top $K$ prediction in terms of ranking scores for every image would not be sufficiently informative. In contrast, logistics and our method applied a more flexible prediction method and were able to achieve more balanced results on precisions and recalls for both datasets.

\pgfplotsset{try min ticks=4}
\begin{figure*}[t]
  \begin{minipage}{.27\linewidth}
      \includegraphics[width=.98\linewidth]{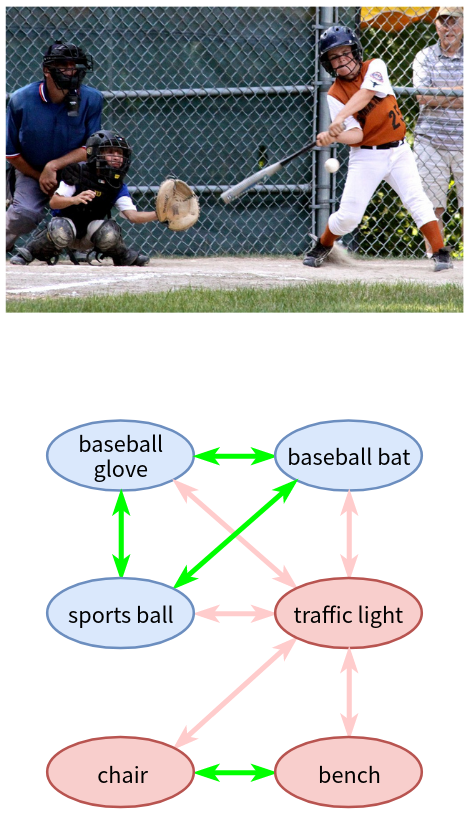}
  \end{minipage}
  \begin{minipage}{.22\linewidth}
      \begin{tikzpicture}
      \begin{axis}[
      		  title style={at={(0.5,0.85)},anchor=south,yshift=1},
              scale only axis,
              title={Baseball glove},
              %symbolic x coords={1,2,3},
              xtick={0,1,2,3,4,5},
              xlabel={$t$},
              ylabel={$p$},
              x label style={
              	at={(1.05,0.58)},
              },
              y label style={
              	at={(0.45,1.15)},
                rotate=-90,
              },
              y tick label style={
        		/pgf/number format/.cd,
            	fixed,
            	fixed zerofill,
            	precision=2,
        		/tikz/.cd
    		  },
              tick label style={
              	font=\scriptsize,
              },
              grid style=dashed,
              width=.7\linewidth,
              height=1.3cm,
          	  no markers,
              ymax=0.9,ymin=0.55,
              xtick pos=left,
				ytick pos=left,
          	enlargelimits=false
          ]
          \addplot coordinates {(0,0.61)(1,0.85)(2,0.853)(3,0.854)(4,0.849)(5,0.836)};
      \end{axis}
      \end{tikzpicture}
      \begin{tikzpicture}
      \begin{axis}[
      		  title style={at={(0.5,0.85)},anchor=south,yshift=1},
              scale only axis,
              title={Sports ball},
              %symbolic x coords={1,2,3},
              xtick={0,1,2,3,4,5},
              xlabel={$t$},
              ylabel={$p$},
              x label style={
              	at={(1.05,0.58)},
              },
              y label style={
              	at={(0.45,1.15)},
                rotate=-90,
              },
              y tick label style={
        		/pgf/number format/.cd,
            	fixed,
            	fixed zerofill,
            	precision=2,
        		/tikz/.cd
    		  },
              tick label style={
              	font=\scriptsize,
              },
              grid style=dashed,
              width=.7\linewidth,
              height=1.3cm,
              ymin=0.45,ymax=0.62,
              xtick pos=left,
				ytick pos=left,
          	  no markers,
          enlargelimits=false
          ]
          \addplot coordinates {(0,0.504)(1,0.59)(2,0.60)(3,0.60)(4,0.578)(5,0.54)};
      \end{axis}
      \end{tikzpicture}
      \begin{tikzpicture}
      \begin{axis}[
      		  title style={at={(0.5,0.85)},anchor=south,yshift=1},
              scale only axis,
              title={Chair},
              %symbolic x coords={1,2,3},
              %xtick=\empty,
              xtick={0,1,2,3,4,5},
              xlabel={$t$},
              ylabel={$p$},
              x label style={
              	at={(1.05,0.58)},
              },
              y label style={
              	at={(0.45,1.15)},
                rotate=-90,
              },
              y tick label style={
        		/pgf/number format/.cd,
            	fixed,
            	fixed zerofill,
            	precision=2,
        		/tikz/.cd
    		  },
              tick label style={
              	font=\scriptsize,
              },
              xtick pos=left,
				ytick pos=left,
              scaled y ticks=false,
              grid style=dashed,
              width=.7\linewidth,
              height=1.3cm,
              ymin=0.015,ymax=0.08,
          	  no markers,
          enlargelimits=false
          ]
          \addplot coordinates {(0,0.06)(1,0.048)(2,0.043)(3,0.042)(4,0.041)(5,0.039)};
      \end{axis}
      \end{tikzpicture}
      \begin{tikzpicture}
      \begin{axis}[
      		  title style={at={(0.5,0.85)},anchor=south,yshift=1},
              scale only axis,
              title={Bench},
              %symbolic x coords={1,2,3},
              %xtick=\empty,
              xtick={0,1,2,3,4,5},
              xlabel={$t$},
              ylabel={$p$},
              x label style={
              	at={(1.05,0.58)},
              },
              y label style={
              	at={(0.45,1.15)},
                rotate=-90,
              },
              y tick label style={
        		/pgf/number format/.cd,
            	fixed,
            	fixed zerofill,
            	precision=2,
        		/tikz/.cd
    		  },
              tick label style={
              	font=\scriptsize,
              },
              xtick pos=left,
				ytick pos=left,
              grid style=dashed,
              width=.7\linewidth,
              height=1.3cm,
              ymin=0.15,ymax=0.21,
          	  no markers,
          enlargelimits=false
          ]
          \addplot coordinates {(0,0.203)(1,0.162)(2,0.164)(3,0.173)(4,0.176)(5,0.169)};
      \end{axis}
      \end{tikzpicture}
  \end{minipage}
  \begin{minipage}{.27\linewidth}
      \includegraphics[width=.98\linewidth]{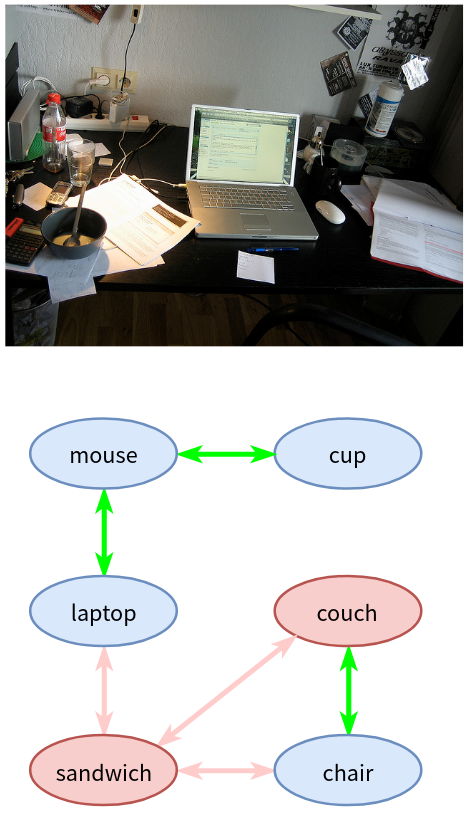}
  \end{minipage}
  \begin{minipage}{.22\linewidth}
      \begin{tikzpicture}
      \begin{axis}[
      		  title style={at={(0.5,0.85)},anchor=south,yshift=1},
              scale only axis,
              title={Mouse},
              xtick={0,1,2,3,4,5},
              xlabel={$t$},
              ylabel={$p$},
              x label style={
              	at={(1.05,0.58)},
              },
              y label style={
              	at={(0.45,1.15)},
                rotate=-90,
              },
              y tick label style={
        		/pgf/number format/.cd,
            	fixed,
            	fixed zerofill,
            	precision=2,
        		/tikz/.cd
    		  },
              tick label style={
              	font=\scriptsize,
              },
              xtick pos=left,
				ytick pos=left,
              grid style=dashed,
              width=.7\linewidth,
              height=1.3cm,
          	  no markers,
          enlargelimits=false
          ]
          \addplot coordinates {(0,0.698)(1,0.827)(2,0.886)(3,0.912)(4,0.928)(5,0.933)};
      \end{axis}
      \end{tikzpicture}
      \begin{tikzpicture}
      \begin{axis}[
      		  title style={at={(0.5,0.85)},anchor=south,yshift=1},
              scale only axis,
              title={Cup},
              %symbolic x coords={1,2,3},
              xtick={0,1,2,3,4,5},
              xlabel={$t$},
              ylabel={$p$},
              x label style={
              	at={(1.05,0.58)},
              },
              y label style={
              	at={(0.45,1.15)},
                rotate=-90,
              },
              y tick label style={
        		/pgf/number format/.cd,
            	fixed,
            	fixed zerofill,
            	precision=2,
        		/tikz/.cd
    		  },
              tick label style={
              	font=\scriptsize,
              },
              xtick pos=left,
				ytick pos=left,
              grid style=dashed,
              width=.7\linewidth,
              height=1.3cm,
          	  no markers,
              ymin=0.39, ymax=0.56,
          enlargelimits=false
          ]
          \addplot coordinates {(0,0.422)(1,0.472)(2,0.506)(3,0.514)(4,0.515)(5,0.537)};
      \end{axis}
      \end{tikzpicture}
      \begin{tikzpicture}
      \begin{axis}[
      		  title style={at={(0.5,0.85)},anchor=south,yshift=1},
              scale only axis,
              title={Chair},
              %symbolic x coords={1,2,3},
              %xtick=\empty,
              xtick={0,1,2,3,4,5},
              xlabel={$t$},
              ylabel={$p$},
              x label style={
              	at={(1.05,0.58)},
              },
              y label style={
              	at={(0.45,1.15)},
                rotate=-90,
              },
              y tick label style={
        		/pgf/number format/.cd,
            	fixed,
            	fixed zerofill,
            	precision=2,
        		/tikz/.cd
    		  },
              tick label style={
              	font=\scriptsize,
              },
              xtick pos=left,
				ytick pos=left,
              scaled y ticks=false,
              grid style=dashed,
              width=.7\linewidth,
              height=1.3cm,
              ymin=0.53,ymax=0.64,
          	  no markers,
          enlargelimits=false
          ]
          \addplot coordinates {(0,0.525)(1,0.552)(2,0.584)(3,0.603)(4,0.610)(5,0.618)};
      \end{axis}
      \end{tikzpicture}
      \begin{tikzpicture}
      \begin{axis}[
      		  title style={at={(0.5,0.85)},anchor=south,yshift=1},
              scale only axis,
              title={Couch},
              %symbolic x coords={1,2,3},
              %xtick=\empty,
              xtick={0,1,2,3,4,5},
              xlabel={$t$},
              ylabel={$p$},
              x label style={
              	at={(1.05,0.58)},
              },
              y label style={
              	at={(0.45,1.15)},
                rotate=-90,
              },
              y tick label style={
        		/pgf/number format/.cd,
            	fixed,
            	fixed zerofill,
            	precision=2,
        		/tikz/.cd
    		  },
              tick label style={
              	font=\scriptsize,
              },
              xtick pos=left,
				ytick pos=left,
              scaled y ticks=false,
              grid style=dashed,
              width=.7\linewidth,
              height=1.3cm,
          	  no markers,
          enlargelimits=false,
              ymin=0.025,ymax=0.065,
          ]
          \addplot coordinates {(0,0.058)(1,0.040)(2,0.041)(3,0.043)(4,0.043)(5,0.044)};
      \end{axis}
      \end{tikzpicture}
  \end{minipage}
\caption{Examples of the constructed knowledge subgraphs and the predicted label probabilities using our proposed method, showing that information propagates across different labels as time step $t$ increases. Note that the blue and red nodes in each subgraph indicate ground truth positive and negative labels, respectively. And, arrows in green or red reflects the corresponding positive or negative relationship.}
\label{fig:fig5}
\end{figure*}
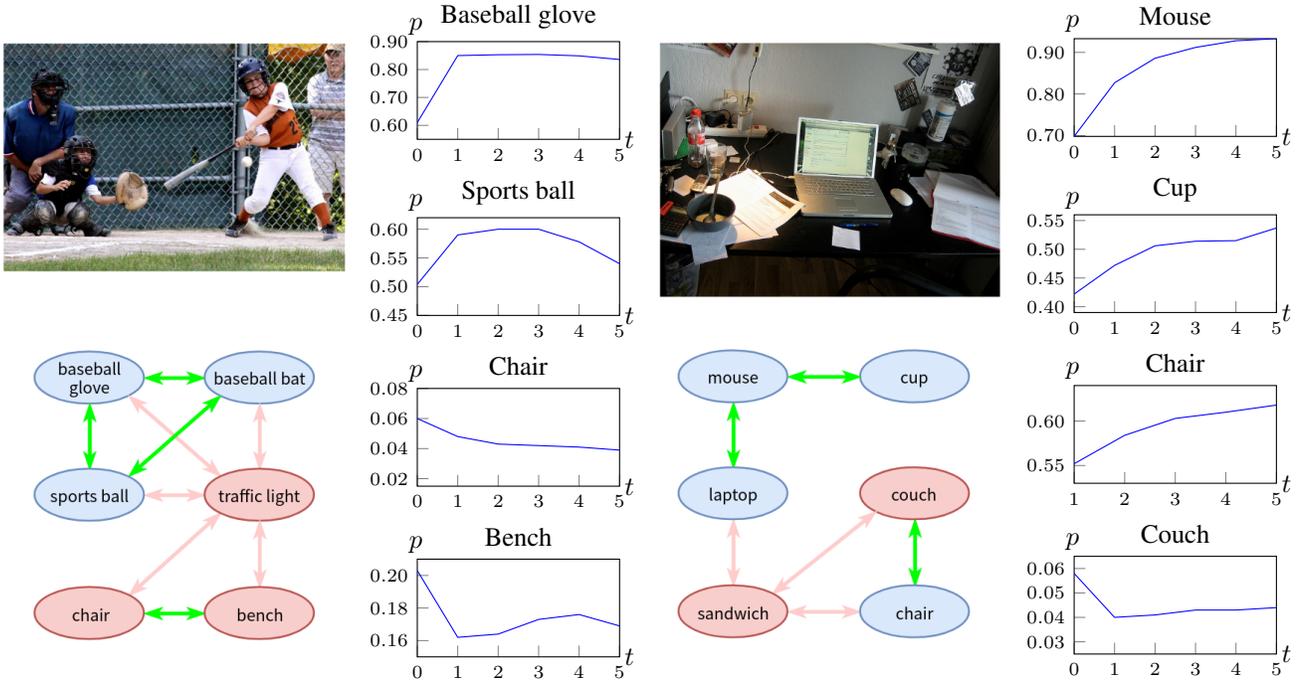

\subsection{ML-ZSL and Generalized ML-ZSL}
We now report our empirical results on multi-label zero-shot learning (ML-ZSL) using the NUS-WIDE dataset. 
In order to perform ML-ZSL, we treat labels in NUS-WIDE 81 as the unseen label set $\mathcal{U}$, while the seen label set $\mathcal{S}$ is derived from NUS-1000 with 75 duplicated ones removed and thus results in 925 label classes.

We take Fast0Tag \cite{Zhang_2016_CVPR} with the same $\mathcal{S}$ and $\mathcal{U}$ as the state-of-the-art ML-ZSL approach for comparisons. We report the results for ML-ZSL with $K=3$ for Fast0Tag. To further verify the effectiveness of the introduced components in our model, we also conduct controlled experiments in which we have a simplified version without updating the belief vectors via the structured knowledge graph (i.e., Ours w/o Prop.). 
In other words, for Ours w/o Prop., we set $T=0$.

Additionally, we consider the challenging task of generalized ML-ZSL task, for which models are trained on seen labels but are required to predict both seen and unseen labels during testing. The experiments are performed on the NUS-WIDE dataset following the ML-ZSL setting, and we report the results of predictions for the $|\mathcal{S}|+|\mathcal{U}|=1006$ labels. 
For Fast0Tag under this setting we report $K=10$, as $K=3$ will result in low recall due to a large number of tags predicted for each image. 

Table~\ref{tb:mlzsl} lists the results for both the ML-ZSL setting and the generalized ML-ZSL setting. 
From this table, we see that our model reported satisfactory performances and performed favorably against Fast0Tag. Also, from the ablation tests, we see that the full version of our model was preferable when applying propagation with the knowledge graph. This confirms the effectiveness of this mechanism introduced in our model.

%It can be seen that our method performs generally better on macro evaluation metrics while attaining comparable results on the micro metrics, showing that less frequent classes can be predicted more accurately with the aid of knowledge graphs information in our model.

\begin{figure}
\centering
  \begin{tikzpicture}
  \begin{axis}[
          title style={
          	at={(0.5,1.0)},
            anchor=south,
            yshift=1,
            font=\normalsize
          },
          scale only axis,
          title={MS-COCO Performance at $t$},
          xtick={0,1,2,3,4,5},
          xlabel={$t$},
          ylabel={F1},
          x label style={
            at={(1.05,0.25)},
          },
          y label style={
            at={(0.15,1.12)},
            rotate=-90,
          },
          y tick label style={
            /pgf/number format/.cd,
            fixed,
            fixed zerofill,
            precision=3,
            /tikz/.cd
          },
          tick label style={
            font=\footnotesize,
          },
              xtick pos=left,
				ytick pos=left,
          scaled y ticks=false,
          grid style=dashed,
          width=.6\linewidth,
          height=.25\linewidth,
          ymin=0.68,ymax=0.695,
          legend pos=south east,
          no markers,
          enlargelimits=false
      ]
      \addplot coordinates {(0,0.6828)(1,0.6890)(2,0.6891)(3,0.6890)(4,0.6892)(5,0.6896)};
  \end{axis}
  \end{tikzpicture}
  
  \vspace{12pt}
  \begin{tikzpicture}
  \begin{axis}[
          title style={
          	at={(0.5,1.0)},
            anchor=south,
            yshift=1,
            font=\normalsize
          },
          scale only axis,
          title={NUS-81 Performance at $t$},
          xtick={0,1,2,3,4,5},
          xlabel={$t$},
          ylabel={F1},
          x label style={
            at={(1.05,0.25)},
          },
          y label style={
            at={(0.15,1.12)},
            rotate=-90,
          },
          y tick label style={
            /pgf/number format/.cd,
            fixed,
            fixed zerofill,
            precision=3,
            /tikz/.cd
          },
          tick label style={
            font=\footnotesize,
          },
              xtick pos=left,
				ytick pos=left,
          scaled y ticks=false,
          grid style=dashed,
          width=.6\linewidth,
          height=.25\linewidth,
          ymin=0.445,ymax=0.46,
          legend pos=south east,
          no markers,
          enlargelimits=false
      ]
      \addplot coordinates {(0,0.4525)(1,0.4544)(2,0.4557)(3,0.4560)(4,0.4566)(5,0.4566)};
  \end{axis}
  \end{tikzpicture}
  \caption{The scores of F1 measure for multi-label classification at different time steps $t$ on MS-COCO and NUS-81.}
  \label{fig:fig6}
\end{figure}
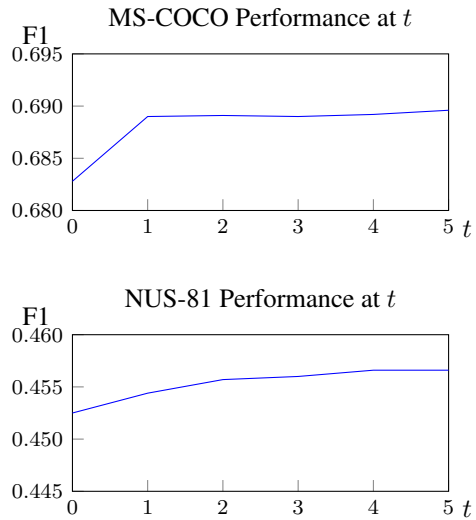
\begin{figure*}
\begin{minipage}{.5\linewidth}

\centering
  \begin{tikzpicture}
  \begin{axis}[
          title style={
          	at={(0.5,1.15)},
            anchor=south,
            yshift=1,
            font=\normalsize
          },
          scale only axis,
          title={NUS-1000 Performance at $t$ (Seen Labels)},
          xtick={0,1,2,3,4,5},
          xlabel={$t$},
          ylabel={F1},
          x label style={
            at={(1.05,0.25)},
          },
          y label style={
            at={(0.2,1.12)},
            rotate=-90,
          },
          y tick label style={
            /pgf/number format/.cd,
            fixed,
            fixed zerofill,
            precision=3,
            /tikz/.cd
          },
          tick label style={
            font=\footnotesize,
          },
              xtick pos=left,
				ytick pos=left,
          scaled y ticks=false,
          grid style=dashed,
          width=.6\linewidth,
          height=.25\linewidth,
          ymin=0.232,ymax=0.234,
          legend pos=south east,
          no markers,
          enlargelimits=false
      ]
      \addplot coordinates {(0,0.2303)(1,0.2327)(2,0.2334)(3,0.2334)(4,0.2330)(5,0.2330)};
      %\addplot coordinates {(0,0.2325)(1,0.2341)(2,0.2345)(3,0.2345)(4,0.2344)(5,0.2341)};
  \end{axis}
  \end{tikzpicture}
\end{minipage}
  \begin{minipage}{.5\linewidth}
\centering
  \begin{tikzpicture}
  \begin{axis}[
          title style={
          	at={(0.5,1.15)},
            anchor=south,
            yshift=1,
            font=\normalsize
          },
          scale only axis,
          title={NUS-1000 Performance at $t$ (Unseen Labels)},
          xtick={0,1,2,3,4,5},
          xlabel={$t$},
          ylabel={F1},
          x label style={
            at={(1.05,0.25)},
          },
          y label style={
            at={(0.2,1.12)},
            rotate=-90,
          },
          y tick label style={
            /pgf/number format/.cd,
            fixed,
            fixed zerofill,
            precision=3,
            /tikz/.cd
          },
          tick label style={
            font=\footnotesize,
          },
              xtick pos=left,
				ytick pos=left,
          scaled y ticks=false,
          grid style=dashed,
          width=.6\linewidth,
          height=.25\linewidth,
          ymin=0.298,ymax=0.305,
          legend pos=south east,
          no markers,
          enlargelimits=false
      ]
      \addplot coordinates {(0,0.3007)(1,0.3039)(2,0.3028)(3,0.3022)(4,0.3016)(5,0.3017)};
      %\addplot coordinates {(0,0.3036)(1,0.3015)(2,0.3006)(3,0.3021)(4,0.3047)(5,0.3061)};
  \end{axis}
  \end{tikzpicture}
  \end{minipage}
  \caption{The scores of F1 measure for seen and unseen labels (i.e., generalized ML-ZSL) at different time steps $t$ on NUS-1000.}
  \label{fig:fig7}
\end{figure*}
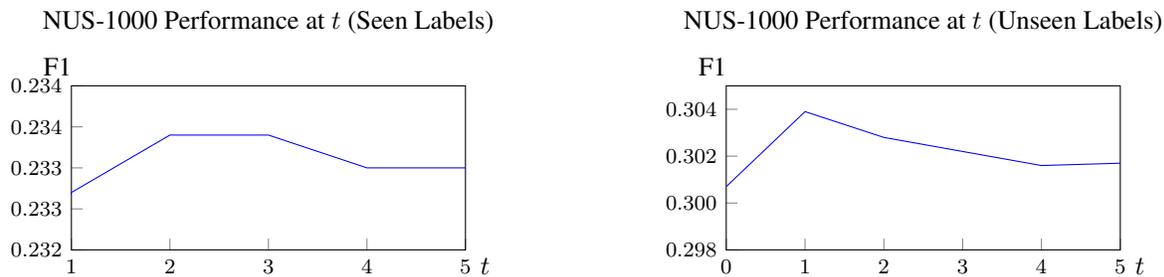

\subsection{Analysis of Propagation Mechanism}
To further evaluate the effectiveness of our method, we visualize the propagation process of our structured knowledge graph in Figure~\ref{fig:fig5}, demonstrating how the information transferred in our constructed graph assists in the prediction process. 
We show the prediction probabilities $p_v^{(t)}$ of several label classes from $t=0$ to $t=5$ for the two examples shown in this figure (both are from MS-COCO). These probabilities are obtained from our multi-label classification model. 
The corresponding knowledge subgraphs are also shown in the figure. From the results, We observe that the first few propagation step affected the prediction probabilities the most, especially for the label nodes that had initial confidence that were closer to the probability threshold. 
Subsequent propagation steps simply further fine-tuned the probabilities for more accurate predictions.

We also made this similar observation when analyzing the performance of our model at different time steps. 
We use the probabilities $p_v^{(t)}$ at time step $t$ instead of time step $T$ to obtain predictions and measure the performance on the testing sets. 
The results for multi-label classification on MS-COCO and NUS-81 for $t=1$ to $t=5$ are shown in Figure~\ref{fig:fig6}.
In Figures~\ref{fig:fig7}, we also observe similar trends for generalized ML-ZSL using NUS-WIDE 1000. In other words, both seen and unseen classes gained from such information propagation across labels, and showed the converged results in a few time steps.
\section{Conclusion}
In this paper, we proposed a unique deep learning framework to approach multi-label learning and multi-label zero-shot learning (ML-ZSL). 
By incorporating structured knowledge graphs into the learning process, our model leverages different relations defined in the constructed knowledge graph, which allow the exploitation of label dependencies between labels for ML-ZSL. This is similar to how humans utilize learned concept dependencies when recognizing seen and unseen objects of interest. In our experiments, we showed that our proposed model was able to produce satisfactory performance on the standard task of multi-label classification, and performed favorably against baseline and state-of-the-art approaches on the challenging problem of ML-ZSL.\\
%In our experiments, we showed that the proposed model was able to improve both multi-label learning and ML-ZSL tasks when comparing to baseline or state-of-the-art deep learning approaches. 

\noindent\textbf{Acknowledgments}
This work was supported in part by the Ministry of Science
and Technology of Taiwan under grant MOST 107-2634-F-002-010.

{\small
\bibliographystyle{ieee}
\bibliography{bibliography/general,bibliography/ml,bibliography/mlzsl,bibliography/graph,bibliography/zsl}

\begin{thebibliography}{10}\itemsep=-1pt

\bibitem{akata2013label}
Z.~Akata, F.~Perronnin, Z.~Harchaoui, and C.~Schmid.
\newblock Label-embedding for attribute-based classification.
\newblock In {\em Proceedings of the IEEE Conference on Computer Vision and
  Pattern Recognition}, pages 819--826, 2013.

\bibitem{akata2015evaluation}
Z.~Akata, S.~Reed, D.~Walter, H.~Lee, and B.~Schiele.
\newblock Evaluation of output embeddings for fine-grained image
  classification.
\newblock In {\em Proceedings of the IEEE Conference on Computer Vision and
  Pattern Recognition}, pages 2927--2936, 2015.

\bibitem{BalasubramanianL12}
K.~Balasubramanian and G.~Lebanon.
\newblock The landmark selection method for multiple output prediction.
\newblock In {\em ICML}. icml.cc / Omnipress, 2012.

\bibitem{changpinyo2016synthesized}
S.~Changpinyo, W.-L. Chao, B.~Gong, and F.~Sha.
\newblock Synthesized classifiers for zero-shot learning.
\newblock In {\em Proceedings of the IEEE Conference on Computer Vision and
  Pattern Recognition}, pages 5327--5336, 2016.

\bibitem{changpinyo2017predicting}
S.~Changpinyo, W.-L. Chao, and F.~Sha.
\newblock Predicting visual exemplars of unseen classes for zero-shot learning.
\newblock In {\em Computer Vision (ICCV), 2017 IEEE International Conference
  on}, pages 3496--3505. IEEE, 2017.

\bibitem{NIPS2012_4561}
Y.-N. Chen and H.-T. Lin.
\newblock Feature-aware label space dimension reduction for multi-label
  classification.
\newblock In F.~Pereira, C.~J.~C. Burges, L.~Bottou, and K.~Q. Weinberger,
  editors, {\em Advances in Neural Information Processing Systems 25}, pages
  1529--1537. Curran Associates, Inc., 2012.

\bibitem{icml2010_DembczynskiCH10}
W.~Cheng, E.~Hüllermeier, and K.~J. Dembczynski.
\newblock Bayes optimal multilabel classification via probabilistic classifier
  chains.
\newblock In J.~Fürnkranz and T.~Joachims, editors, {\em Proceedings of the
  27th International Conference on Machine Learning (ICML-10)}, pages 279--286.
  Omnipress, 2010.

\bibitem{04d503dffe9f4dcfbacf886d1dee56fb}
K.~Cho, B.~{van Merrienboer}, D.~Bahdanau, and Y.~Bengio.
\newblock On the properties of neural machine translation: Encoder-decoder
  approaches.
\newblock In {\em Eighth Workshop on Syntax, Semantics and Structure in
  Statistical Translation (SSST-8), 2014}, 2014.

\bibitem{chua2009nus}
T.-S. Chua, J.~Tang, R.~Hong, H.~Li, Z.~Luo, and Y.~Zheng.
\newblock Nus-wide: a real-world web image database from national university of
  singapore.
\newblock In {\em Proceedings of the ACM international conference on image and
  video retrieval}, page~48. ACM, 2009.

\bibitem{deng2014large}
J.~Deng, N.~Ding, Y.~Jia, A.~Frome, K.~Murphy, S.~Bengio, Y.~Li, H.~Neven, and
  H.~Adam.
\newblock Large-scale object classification using label relation graphs.
\newblock In {\em European Conference on Computer Vision}, pages 48--64.
  Springer, 2014.

\bibitem{deng2009imagenet}
J.~Deng, W.~Dong, R.~Socher, L.-J. Li, K.~Li, and L.~Fei-Fei.
\newblock Imagenet: A large-scale hierarchical image database.
\newblock In {\em Computer Vision and Pattern Recognition, 2009. CVPR 2009.
  IEEE Conference on}, pages 248--255. IEEE, 2009.

\bibitem{deng2016structure}
Z.~Deng, A.~Vahdat, H.~Hu, and G.~Mori.
\newblock Structure inference machines: Recurrent neural networks for analyzing
  relations in group activity recognition.
\newblock In {\em Proceedings of the IEEE Conference on Computer Vision and
  Pattern Recognition}, pages 4772--4781, 2016.

\bibitem{ding2015probabilistic}
N.~Ding, J.~Deng, K.~P. Murphy, and H.~Neven.
\newblock Probabilistic label relation graphs with ising models.
\newblock In {\em Proceedings of the IEEE International Conference on Computer
  Vision}, pages 1161--1169, 2015.

\bibitem{FarhadiEHF09}
A.~Farhadi, I.~Endres, D.~Hoiem, and D.~A. Forsyth.
\newblock Describing objects by their attributes.
\newblock In {\em CVPR}, pages 1778--1785. IEEE Computer Society, 2009.

\bibitem{frome2013devise}
A.~Frome, G.~S. Corrado, J.~Shlens, S.~Bengio, J.~Dean, T.~Mikolov, et~al.
\newblock {DeViSE}: A deep visual-semantic embedding model.
\newblock In {\em Advances in neural information processing systems}, pages
  2121--2129, 2013.

\bibitem{embedding2014ECCV}
Y.~Fu, Y.~Yang, T.~M. Hospedales, T.~Xiang, and S.~Gong.
\newblock Transductive multi-label zero-shot learning.
\newblock In {\em BMVC}, 2014.

\bibitem{Gaure_2017_UAI}
A.~Gaure, A.~Gupta, V.~K. Verma, and P.~Rai.
\newblock A probabilistic framework for zero-shot multi-label learning.
\newblock In {\em The Conference on Uncertainty in Artificial Intelligence
  (UAI)}, 2017.

\bibitem{DBLP:journals/corr/GongJLTI13}
Y.~Gong, Y.~Jia, T.~Leung, A.~Toshev, and S.~Ioffe.
\newblock Deep convolutional ranking for multilabel image annotation.
\newblock {\em CoRR}, abs/1312.4894, 2013.

\bibitem{He_2016_CVPR}
K.~He, X.~Zhang, S.~Ren, and J.~Sun.
\newblock Deep residual learning for image recognition.
\newblock In {\em The IEEE Conference on Computer Vision and Pattern
  Recognition (CVPR)}, June 2016.

\bibitem{hochreiter1997long}
S.~Hochreiter and J.~Schmidhuber.
\newblock Long short-term memory.
\newblock {\em Neural computation}, 9(8):1735--1780, 1997.

\bibitem{hu2016learning}
H.~Hu, G.-T. Zhou, Z.~Deng, Z.~Liao, and G.~Mori.
\newblock Learning structured inference neural networks with label relations.
\newblock In {\em Proceedings of the IEEE Conference on Computer Vision and
  Pattern Recognition}, pages 2960--2968, 2016.

\bibitem{NIPS2014_5290}
D.~Jayaraman and K.~Grauman.
\newblock Zero-shot recognition with unreliable attributes.
\newblock In Z.~Ghahramani, M.~Welling, C.~Cortes, N.~Lawrence, and
  K.~Weinberger, editors, {\em Advances in Neural Information Processing
  Systems 27}, pages 3464--3472. Curran Associates, Inc., 2014.

\bibitem{krizhevsky2012imagenet}
A.~Krizhevsky, I.~Sutskever, and G.~E. Hinton.
\newblock Imagenet classification with deep convolutional neural networks.
\newblock In {\em Advances in neural information processing systems}, pages
  1097--1105, 2012.

\bibitem{lampert2009learning}
C.~H. Lampert, H.~Nickisch, and S.~Harmeling.
\newblock Learning to detect unseen object classes by between-class attribute
  transfer.
\newblock In {\em Computer Vision and Pattern Recognition, 2009. CVPR 2009.
  IEEE Conference on}, pages 951--958. IEEE, 2009.

\bibitem{Lampert:2014:ACZ:2587733.2587824}
C.~H. Lampert, H.~Nickisch, and S.~Harmeling.
\newblock Attribute-based classification for zero-shot visual object
  categorization.
\newblock {\em IEEE Trans. Pattern Anal. Mach. Intell.}, 36(3):453--465, Mar.
  2014.

\bibitem{ba15iccv}
J.~Lei~Ba, K.~Swersky, S.~Fidler, and R.~salakhutdinov.
\newblock Predicting deep zero-shot convolutional neural networks using textual
  descriptions.
\newblock In {\em The IEEE International Conference on Computer Vision (ICCV)},
  December 2015.

\bibitem{DBLP:journals/corr/LiTBZ15}
Y.~Li, D.~Tarlow, M.~Brockschmidt, and R.~S. Zemel.
\newblock Gated graph sequence neural networks.
\newblock {\em CoRR}, abs/1511.05493, 2015.

\bibitem{502}
T.-Y. Lin, M.~Maire, S.~Belongie, J.~Hays, P.~Perona, D.~Ramanan, P.~Dollár,
  and C.~L. Zitnick.
\newblock Microsoft coco: Common objects in context.
\newblock In {\em European Conference on Computer Vision (ECCV)}, Zürich,
  2014.
\newblock Oral.

\bibitem{lin2014multi}
Z.~Lin, G.~Ding, M.~Hu, and J.~Wang.
\newblock Multi-label classification via feature-aware implicit label space
  encoding.
\newblock In {\em International Conference on Machine Learning}, pages
  325--333, 2014.

\bibitem{Marino_2017_CVPR}
K.~Marino, R.~Salakhutdinov, and A.~Gupta.
\newblock The more you know: Using knowledge graphs for image classification.
\newblock In {\em The IEEE Conference on Computer Vision and Pattern
  Recognition (CVPR)}, July 2017.

\bibitem{Mensink_2014_CVPR}
T.~Mensink, E.~Gavves, and C.~G. Snoek.
\newblock Costa: Co-occurrence statistics for zero-shot classification.
\newblock In {\em The IEEE Conference on Computer Vision and Pattern
  Recognition (CVPR)}, June 2014.

\bibitem{NIPS2013_5021}
T.~Mikolov, I.~Sutskever, K.~Chen, G.~S. Corrado, and J.~Dean.
\newblock Distributed representations of words and phrases and their
  compositionality.
\newblock In C.~J.~C. Burges, L.~Bottou, M.~Welling, Z.~Ghahramani, and K.~Q.
  Weinberger, editors, {\em Advances in Neural Information Processing Systems
  26}, pages 3111--3119. Curran Associates, Inc., 2013.

\bibitem{Miller:1995:WLD:219717.219748}
G.~A. Miller.
\newblock Wordnet: A lexical database for english.
\newblock {\em Commun. ACM}, 38(11):39--41, Nov. 1995.

\bibitem{DBLP:journals/corr/NamKGF13}
J.~Nam, J.~Kim, I.~Gurevych, and J.~F{\"{u}}rnkranz.
\newblock Large-scale multi-label text classification - revisiting neural
  networks.
\newblock {\em CoRR}, abs/1312.5419, 2013.

\bibitem{norouzi14iclr}
M.~Norouzi, T.~Mikolov, S.~Bengio, Y.~Singer, J.~Shlens, A.~Frome, G.~Corrado,
  and J.~Dean.
\newblock Zero-shot learning by convex combination of semantic embeddings.
\newblock In {\em International Conference on Learning Representations (ICLR)},
  2014.

\bibitem{NIPS2009_0395}
M.~Palatucci, D.~Pomerleau, G.~E. Hinton, and T.~M. Mitchell.
\newblock Zero-shot learning with semantic output codes.
\newblock In Y.~Bengio, D.~Schuurmans, J.~Lafferty, C.~Williams, and
  A.~Culotta, editors, {\em Advances in Neural Information Processing Systems
  22}, pages 1410--1418. 2009.

\bibitem{pennington2014glove}
J.~Pennington, R.~Socher, and C.~D. Manning.
\newblock Glove: Global vectors for word representation.
\newblock In {\em Empirical Methods in Natural Language Processing (EMNLP)},
  pages 1532--1543, 2014.

\bibitem{Read2011}
J.~Read, B.~Pfahringer, G.~Holmes, and E.~Frank.
\newblock Classifier chains for multi-label classification.
\newblock {\em Machine Learning}, 85(3):333, Jun 2011.

\bibitem{Ren_etal_BMVC_17}
Z.~Ren, H.~Jin, Z.~Lin, C.~Fang, and A.~Yuille.
\newblock Multiple instance visual-semantic embedding.
\newblock In {\em Proceeding of the British Machine Vision Conference (BMVC)},
  2017.

\bibitem{scarselli2009graph}
F.~Scarselli, M.~Gori, A.~C. Tsoi, M.~Hagenbuchner, and G.~Monfardini.
\newblock The graph neural network model.
\newblock {\em IEEE Transactions on Neural Networks}, 20(1):61--80, 2009.

\bibitem{schuster1997bidirectional}
M.~Schuster and K.~K. Paliwal.
\newblock Bidirectional recurrent neural networks.
\newblock {\em IEEE Transactions on Signal Processing}, 45(11):2673--2681,
  1997.

\bibitem{socher2013zero}
R.~Socher, M.~Ganjoo, C.~D. Manning, and A.~Ng.
\newblock Zero-shot learning through cross-modal transfer.
\newblock In {\em Advances in neural information processing systems}, pages
  935--943, 2013.

\bibitem{doi:10.1162/NECO_a_00320}
F.~Tai and H.-T. Lin.
\newblock Multilabel classification with principal label space transformation.
\newblock {\em Neural Computation}, 24(9):2508--2542, 2012.
\newblock PMID: 22594831.

\bibitem{Tsoumakas07multi-labelclassification:}
G.~Tsoumakas and I.~Katakis.
\newblock Multi-label classification: An overview.
\newblock {\em Int J Data Warehousing and Mining}, 2007:1--13, 2007.

\bibitem{Wang_2016_CVPR}
J.~Wang, Y.~Yang, J.~Mao, Z.~Huang, C.~Huang, and W.~Xu.
\newblock {CNN-RNN:} a unified framework for multi-label image classification.
\newblock In {\em The IEEE Conference on Computer Vision and Pattern
  Recognition (CVPR)}, June 2016.

\bibitem{DBLP:journals/corr/WeiXHNDZY14}
Y.~Wei, W.~Xia, J.~Huang, B.~Ni, J.~Dong, Y.~Zhao, and S.~Yan.
\newblock {CNN:} single-label to multi-label.
\newblock {\em CoRR}, abs/1406.5726, 2014.

\bibitem{weston2011wsabie}
J.~Weston, S.~Bengio, and N.~Usunier.
\newblock Wsabie: Scaling up to large vocabulary image annotation.
\newblock In {\em IJCAI}, volume~11, pages 2764--2770, 2011.

\bibitem{wu1994verbs}
Z.~Wu and M.~Palmer.
\newblock Verbs semantics and lexical selection.
\newblock In {\em Proceedings of the 32nd annual meeting on Association for
  Computational Linguistics}, pages 133--138. Association for Computational
  Linguistics, 1994.

\bibitem{yeh2017learning}
C.-K. Yeh, W.-C. Wu, W.-J. Ko, and Y.-C.~F. Wang.
\newblock Learning deep latent space for multi-label classification.
\newblock In {\em AAAI}, pages 2838--2844, 2017.

\bibitem{1683770}
M.-L. Zhang and Z.-H. Zhou.
\newblock Multilabel neural networks with applications to functional genomics
  and text categorization.
\newblock {\em IEEE Transactions on Knowledge and Data Engineering},
  18(10):1338--1351, Oct 2006.

\bibitem{Zhang_2016_CVPR}
Y.~Zhang, B.~Gong, and M.~Shah.
\newblock Fast zero-shot image tagging.
\newblock In {\em The IEEE Conference on Computer Vision and Pattern
  Recognition (CVPR)}, June 2016.

\end{thebibliography}
}

\end{document}